\documentclass{article}

 \usepackage[nonatbib,preprint]{nips_2018}

\usepackage[utf8]{inputenc} 
\usepackage[T1]{fontenc}    
\usepackage{hyperref}       
\usepackage{url}            
\usepackage{booktabs}       
\usepackage{amsfonts}       
\usepackage{nicefrac}       
\usepackage{microtype}      
\usepackage{color}
\usepackage{subfigure}
\usepackage{graphicx}

\usepackage{amsmath}
\usepackage{mathtools}
\DeclarePairedDelimiterX{\ip}[1]{\langle}{\rangle}{#1}

\DeclareMathOperator*{\argmin}{arg\,min}
\DeclareMathOperator{\Log}{Log}
\DeclareMathOperator{\Exp}{Exp}
\DeclareMathOperator*{\Symp}{Sym_+}
\DeclareMathOperator{\Id}{Id}

\title{Latent Space Non-Linear Statistics}

\author{
 Line K\"uhnel \\
  University of Copenhagen\\
  Denmark\\
  \texttt{kuhnel@di.ku.dk} \\
\And 
  Tom Fletcher    \\
  University of Utah\\
  USA\\
  \texttt{fletcher@sci.utah.edu} \\
\And 
  Sarang Joshi    \\
  University of Utah\\
  USA\\
  \texttt{sjoshi@sci.utah.edu} \\
\And 
  Stefan Sommer    \\
  University of Copenhagen\\
  Denmark\\
  \texttt{sommer@di.ku.dk} \\
}

\begin{document}

\maketitle

\begin{abstract}
 Given data, deep generative models, such as variational autoencoders (VAE) and generative adversarial networks (GAN), train a lower dimensional latent representation of the data space. The linear Euclidean geometry of data space pulls back to a nonlinear Riemannian geometry on the latent space. The latent space thus provides a low-dimensional nonlinear representation of data and classical linear statistical techniques are no longer applicable. In this paper we show how statistics of data in their latent space representation can be performed using techniques from the field of nonlinear manifold statistics. Nonlinear manifold statistics provide generalizations of Euclidean statistical notions including means, principal component analysis, and maximum likelihood fits of parametric probability distributions. We develop new techniques for maximum likelihood inference in latent space, and adress the computational complexity of using geometric algorithms with high-dimensional data by training a separate neural network to approximate the Riemannian metric and cometric tensor capturing the shape of the learned data manifold.
\end{abstract}

\section{Introduction}
The Riemannian geometry of latent models, provided by deep generative models, have recently been explored in \cite{shao_riemannian_2017,chen_metrics_2017,arvanitidis_latent_2017}. The mapping $f:Z\rightarrow X$, from latent space $Z$ to the data space $X$, constitutes an embedding of $Z$ into $X$ under mild assumptions on the network architecture. This allows the image $f(Z)$ to inherit the Riemannian metric and hence the geometry from the Euclidean ambient space $X$. Equivalently, the metric structure of $X$ pulls back via $f$ to a nonlinear Riemannian structure on $Z$. The above papers explore aspects of this geometry including numerical schemes for geodesic integration, parallel transport, Fr\'echet mean estimation, simulation of Brownian motion, and interpolation. With this paper, we wish to focus on performing subsequent statistics after learning the latent representation and the embedding $f$. We aim at using the constructions, tools and methods from nonlinear statistics \cite{pennec_intrinsic_2006} to perform statistical analysis of data in the latent representation. 

Deep generative models are excellent tools for learning the intrinsic geometry of a low-dimensional data manifold $f(Z)$, subspace of the data space $X$. When the major modes of data variation are of low intrinsic dimensionality, statistical analyses exploiting the lower dimensionality can be more efficient than performing statistics directly in the high-dimensional data space. By performing statistics in lower-dimensional manifolds learned with deep generative models, we simultaneously adapt the statistics to the intrinsic geometry of the data manifold, exploit the compact representation, and avoid unnecessary dimensions in the high-dimensional space $X$ affecting the statistical analysis.

Exemplified on two datasets, synthetic data on the sphere $\mathbb S^2$ for visualization and the MNIST digits dataset, we show how statistical procedures such as principal component analysis can be performed on the latent space. We will subsequently define and infer parameters of geometric distributions allowing the definition and inference of maximum likelihood estimates via simulation of diffusion processes. Both VAEs and GANs themselves learn distributions representing the input training data. The aim is to perform nonlinear statistical analyses for data independent of the training data and with a different distribution, but which are elements of the same low-dimensional manifold of the data space. The latent representation can in this way be learned unsupervised from large numbers of unlabeled training samples while subsequent low-sample size statistics can be performed using the low-dimensional latent representation. This setting occurs for example in medical imaging where brain MR scans are abundant while controlled disease progression studies are of a much smaller sample size. The approach resembles the common task of using principal component analysis to represent data in the span of fewer principal eigenvectors, with the important difference that in the present case a nonlinear manifold is learned using deep generative models instead of standard linear subspace approximation.

The field of nonlinear statistics provide generalizations of statistical constructions and tools from linear Euclidean vector spaces to Riemannian manifolds. Such constructs, e.g. the mean value, often have many equivalent definitions in Euclidean space. However, nonlinearity and curvature generally break this equivalence leading to a plethora of different generalizations. For this reason, we here focus on a subset of selected methods to exemplify the use of nonlinear statistics tools in the latent space setting: Principal component analysis on manifolds with the principal geodesic analysis (PGA, \cite{fletcher_principal_2004-1}) and inference of maximum likelihood means from intrinsic diffusion processes \cite{sommer_modelling_2017}.

The learned manifold defines a Riemannian metric on the latent representation, still the often high dimensionality of the data manifold makes evaluation of the metric computationally costly. This is severely amplified when calculating higher-order derivatives needed for geometric concepts such as the curvature tensor and the Christoffel symbols that are crucial for numerical integration of geodesics and simulation of sample paths for Brownian motions. We present a new method for handling the computational complexity of evaluating the metric by training a second neural network to approximate the local metric tensor of the latent space thereby achieving a massive speed up in the implementation of the geometric, and nonlinear statistical algorithms.

The paper thus presents the following contributions:
\begin{enumerate}
  \item we couple tools from nonlinear statistics with deep end-to-end differentiable generative models for analyzing data using a pre-trained low-dimensional latent representation,
  \item we show how an additional neural network can be trained to learn the metric tensor and thereby greatly speed up the computations needed for the nonlinear statistics algorithms, 
  \item we develop a method for maximum likelihood estimation of diffusion processes in the latent geometry and use this to estimate ML means from Riemannian Brownian motions.
\end{enumerate}

We show examples of the presented methods on latent geometries learned from synthetic data in $\mathbb R^3$ and on the MNIST dataset. The statistical computations are implemented in the Theano Geometry package \cite{kuhnel_differential_2017} that using the automatic differentiation features of Theano \cite{the_theano_development_team_theano:_2016} allows for easy and concise expression of differential geometry concepts.

The paper starts with a brief description on latent space geometry based on the papers \cite{shao_riemannian_2017,chen_metrics_2017,arvanitidis_latent_2017}. We then discuss definition of mean values in the nonlinear latent geometry and use of the principal geodesic analysis (PGA) procedure before developing a scheme for maximum likelihood estimation of parameters with Riemannian Brownian motion using a diffusion bridge sampling scheme. We end the paper with experiments.

\section{Latent Space Geometry}
\begin{figure}[htpb]
    \centering
    \subfigure{\includegraphics[width=.37\columnwidth,trim=140 100 100 110,clip=true]{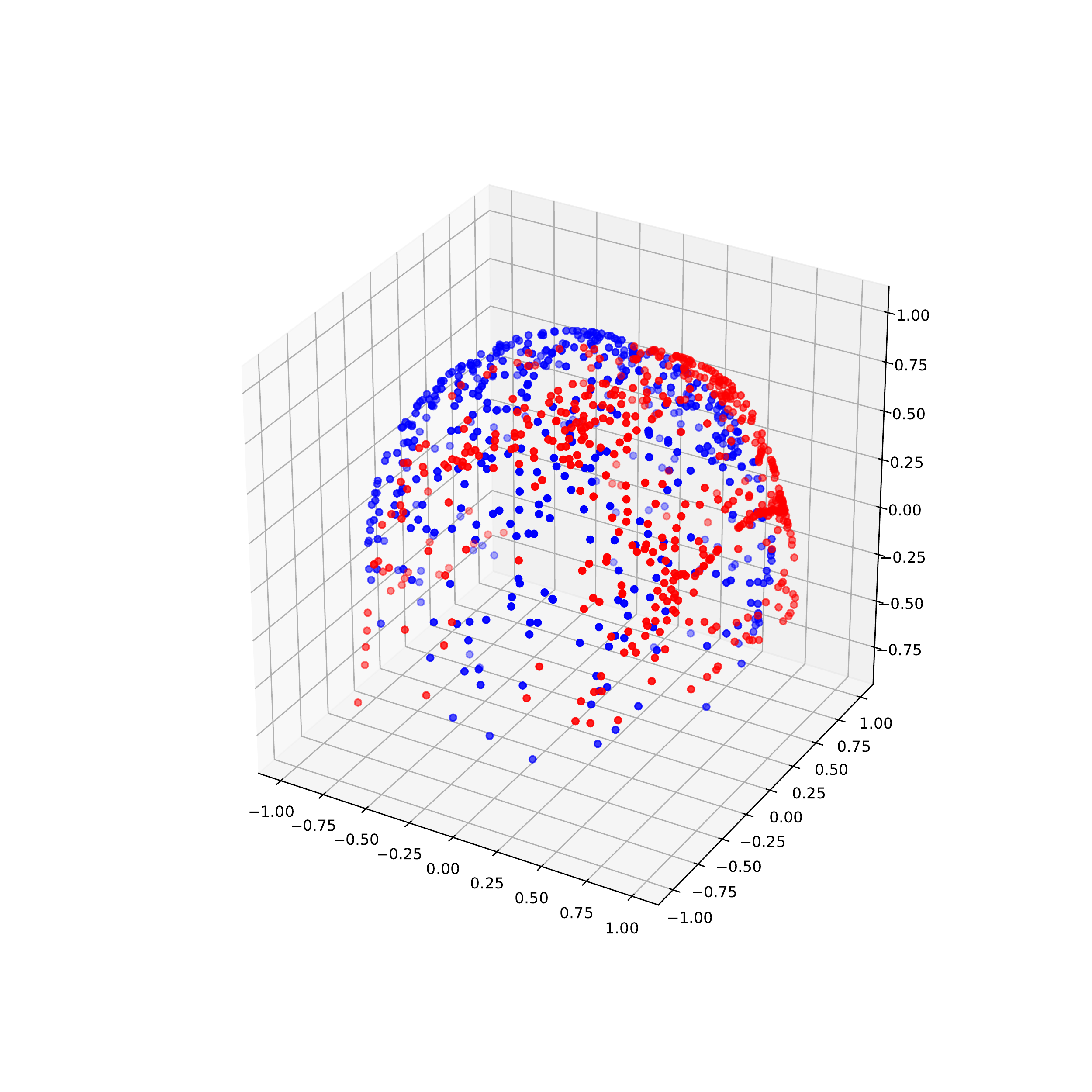}}\hspace{1cm}
    \subfigure{\includegraphics[width=.37\columnwidth,trim=170 120 130 150,clip=true]{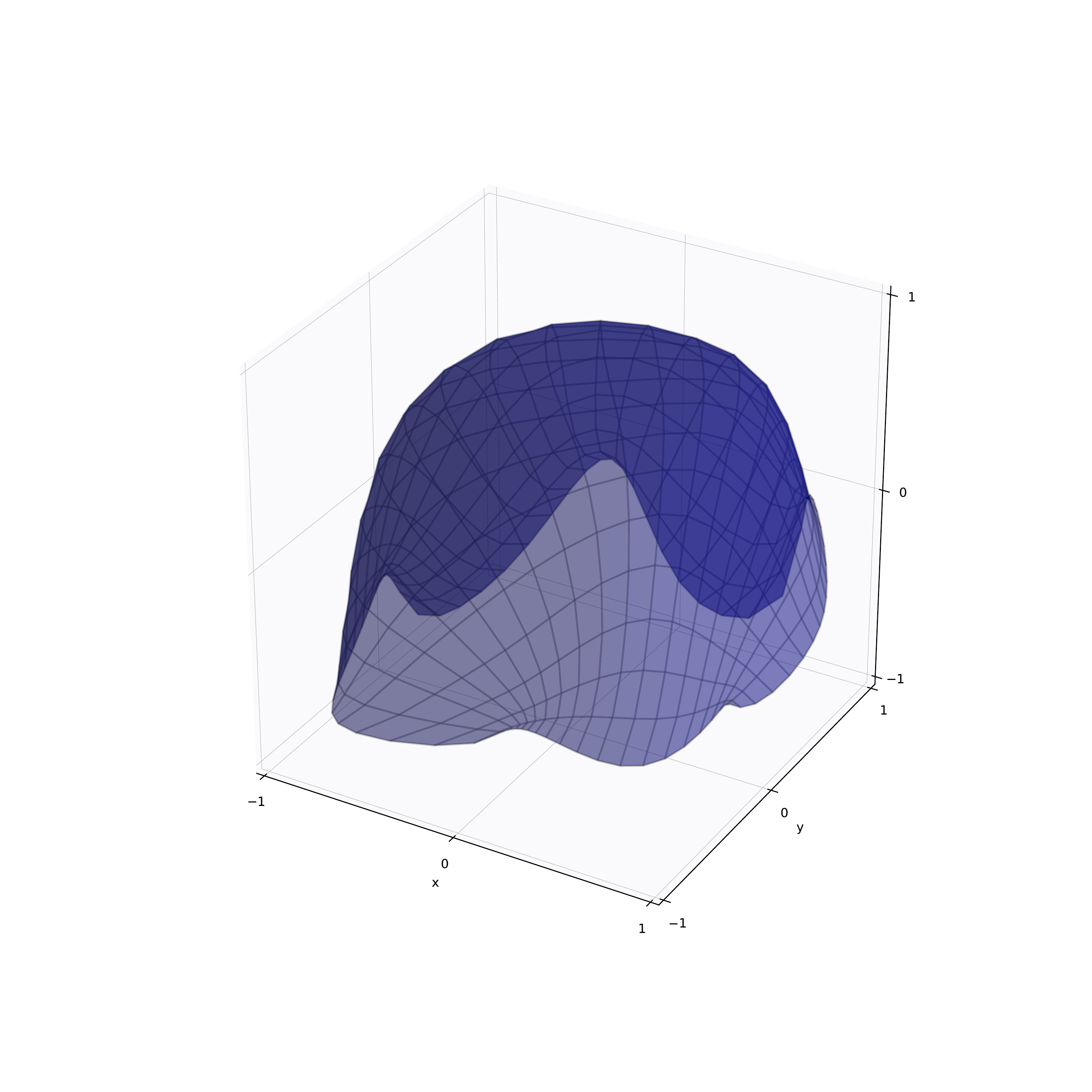}}
    \caption{
     (left) Samples from the data distribution (blue) with corresponding predictions from the VAE (red). (right) The trained manifold.
    }
    \label{fig:S2}
\end{figure}

Deep generative models such as generative adversarial networks (GANs, \cite{goodfellow_generative_2014}) and autoencoders/variational autoencoders (VAEs, \cite{bengio_learning_2009}) learn mappings from a latent space $Z$ to the data space $X$. In the VAE case, the decoder mapping $f\colon Z\to X$ describes the mean of the data distribution, $P(X|z) = \mathcal{N}(X\mid f(z), \sigma(z)^2I)$, and is complemented by an encoder $h\colon X\to Z$. 
Both $Z$ and $X$ are Euclidean spaces, with dimension $d$ and $n$ respectively and generally $d\ll n$. When the pushforward $f_*$, and the differential $df$ of $f$, is of rank $d$ for any point $z$, the image $f(Z)$ in $X$ is an embedded differentiable manifold of dimension $d$. We denote this manifold by $M$. Generally for deep models, $f$ is nonlinear making $M$ a nonlinear manifold. An example of a trained manifold with a VAE is shown in Figure~\ref{fig:S2}. Here we simulate synthetic data on the sphere $\mathbb S^2$ by the transition distribution of a Riemannian Brownian motion starting at the north pole. The learned submanifold approximates $\mathbb{S}^2$ on the northern hemisphere containing the greatest concentration of samples.

The learned manifold $M$ inherits differential and geometric structure from $X$. In particular, the standard Euclidean inner product restricts to tangent spaces $T_xM$ for $x\in M$ to give a Riemannian metric $g$ on $M$, i.e. for $v,w\in T_xM$, $g(v,w)=\ip{v,w}=v^Tw$. Locally, we invert $f$ to obtain charts on $M$, and get the standard expression $g_{ij}(z)=\ip{\partial_{z_i}f,\partial_{z_j}f}$ for the metric tensor in $Z$ coordinates. Using Jacobian matrix $Jf=(\partial_{z_i}f^j)^i_j$, the matrix expression of $g(z)$ is $g(z)=(Jf(z))^TJf(z)$. The metric tensor on $Z$ can be seen as the pullback $f^*g$ of the Riemannian metric on $X$.

The geometry of latent spaces was explored in \cite{shao_riemannian_2017}. In addition to setting up the geometric foundation, the paper developed efficient algorithms for geodesic integration, parallel transport, and Fr\'echet mean estimation on the latent space. The algorithms make particular use of the encoder function $h\colon X\to Z$ that is trained as part of the VAEs. Instead of explicitly computing Christoffel symbols for geodesic integration, the presence of $h$ allows steps of the integration algorithm to be taken in $X$ and then subsequently mapped back to $Z$. Avoiding computation of Christoffel symbols significantly increases execution speed, a critical improvement for the heavy computations involved with the typically high dimensions of $X$. \cite{chen_metrics_2017} provides additional views on the latent geometry and interpolation examples on the MNIST dataset and robotic arm movements. \cite{arvanitidis_latent_2017} includes the $z$-variability of the variance $\sigma(z)$ of VAEs resulting in the inclusion of the Jacobian of $\sigma$ in the expected metric. The paper in addition explores random walks in the latent geometry and how to enable meaningful extrapolation of the latent representation beyond the training data.

\subsection{Latent Data Representations}
\label{sec:latent}
Given sampled data $y_1,\ldots,y_N$ in $X$, the aim is here to perform statistics on the data after mapping to the low-dimensional latent space $Z$. Note that the mapping $f$ can thus be trained unsupervised and afterwards used to perform statistics on new data in the low-dimensional representation. Therefore, the data $y_1,\ldots,y_N$ are generally different from the training data used to train $f$. In particular, $N$ can be much lower than the size of the training set.

For VAEs, the mapping of $y_i$ to corresponding points in the latter representation $z_i$ is directly available from the encoder function $h$, i.e. $z_i=h(y_i)$ 
In more general settings where $h$ is not present, we need to construct $z_i$ from $y_i$. A natural approach is to define $z_i$ from the optimization problem
\begin{equation}
  z_i
  =
  \argmin_{z\in Z}
  \|f(z)-y_i\|^2
  \ .
  \label{eq:zi}
\end{equation}
This can be seen as a projection from $X$ to $M$ using the Euclidean distance in $X$.

\subsection{Geodesics and Brownian Motions}
The pullback metric $f^*g$ on $Z$ defines geometric concepts such as geodesics, exponential and logarithm map, and Riemannian Brownian motions on $Z$. Using $f$, each of these definitions is equivalently expressed on $M$ viewing it as a submanifold of $X$ with inherited metric. Given $z\in Z$ and $v\in T_z Z$, the exponential map $\Exp_z\colon T_z Z\to Z$ is defined as the geodesic $\gamma^v_t$ at time $t=1$ with starting point $z$ and initial velocity $v$, i.e. $\text{Exp}_z(v) = \gamma^v_1$. The logarithm map $\Log\colon Z\times Z\to TZ$ is the local inverse of $\Exp$: Given two points $z_1,z_2\in Z$, $\text{Log}_{z_1}(z_2)$, returns the tangent vector $v\in T_{z_1} Z$ defining the minimizing geodesic between $z_1$ and $z_2$. The Riemannian metric defines the geodesic distance expressed from the logarithm map by $d(z_1,z_2) = \|\text{Log}_{z_1}(z_2)\|_g^2$. Using $Z$ as coordinates for $M$ by local inverses of $f$, the Riemannian Brownian motions on $Z$, and equivalently on $M$, is defined by the coordinate expression
\begin{equation}
  dz^j_t = g(z)^{kl}\Gamma^j_{kl}dt + \sqrt{(g(z))^{-1}}^jdB_{t,j},
\label{eq:brown_motion}
\end{equation}
where $\Gamma^j_{kl}$ denotes Christoffel symbols, $g^{-1}$ the cometric, i.e. the inverse of the metric tensor $g$, $B_t$ a standard Brownian motion in $\mathbb R^d$, and where Einstein notation is used for index summation.

\section{Computational Representation}
While metric computation is easily expressed using automatic differentiation to compute the Jacobian $Jf$ of the embedding map $f$, the high dimensionality of the data space has a computational cost when evaluating the metric. This is particularly emphasized when computing higher-order differential concepts such as Christoffel symbols, used for geodesic integration, curvature, and Brownian motion simulation, due to multiple derivatives and metric inverse computations involved. For integration of geodesics and Brownian motion, one elegant way to avoid the computation of Christoffel symbols is to take each step of the integration in the ambient data space of $M$ and map the result back to the latent space using the encoder mapping $h$ \cite{shao_riemannian_2017}. This requires $h$ to be close to the inverse of $f$ restricted to $M$, and limits the method to VAEs where $h$ is trained along with the decoder, $f$.

We here propose an additional way to allow efficient computations without using the encoder map $h$. The approach therefore works for both GANs and VAEs. The latent space $Z$ is of low dimension, and the only entity needed for encoding the geometry is the metric $g:Z\to\Symp(d)$ that to each $z$ assigns a positive symmetric $d\times d$ matrix. $\Symp(d)$ has dimension $d(d+1)/2$. The high dimensionality of the data space thus does not appear directly when defining the geometry, and $X$ is only used for the actual computation of $g(z)$. We therefore train a second neural network $\tilde{g}$ to act as a function approximator for $g$, i.e. we train $\tilde{g}$ to produce an element of $\Symp(d)$ that is close to $g(z)$ for each $z$. Notice that this network does not evaluate a Jacobian matrix when computing $g(z)$ and no derivatives are hence needed for evaluating the metric. Because of this and due to both input and output space of the network being of low dimensions, $d$ and $d(d+1)/2$ respectively, the computational effort of evaluating $\tilde{g}$ and Christoffel symbols computed from $\tilde{g}$ is orders of magnitude faster than evaluating $g$ directly when the dimensionality $n$ of $X$ is high compared to $d$: Integration of the geodesic equation with 100 timesteps in the MNIST case presented later takes $\approx 30$ s., when computing the metric from $Jf$, compared to $\approx 30$ ms. using the second neural network to predict $g$.

Inverting $\tilde{g}(z)$ is sensitive to the approximation of $g$ provided by $\tilde{g}$. The cometric tensor $g^{-1}$ is therefore more sensitive to the approximation when computed from $\tilde{g}$ than from $g$ itself. This is emphasized when $g(z)$ has small eigenvalues. As a solution, we let the second neural network predict both the metric $g(z)$ and cometric $g(z)^{-1}$. Defining the loss function for training the network, we balance the norm between predicted matrices $\tilde{g}$ and $\tilde{g}^{-1}$. In addition, we ensure that the predicted $\tilde{g}$ and $\tilde{g}^{-1}$ are close to being actual inverses. These observations are expressed in the loss function
\begin{align}
\label{eq:approximator_loss}
    \text{loss}&_{g,g^{-1}\text{-approximator}}(g_{\text{true}},g^{-1}_{\text{true}},g_{\text{predicted}},g^{-1}_{\text{predicted}})
    \\&\qquad
    =
    \|g_{\text{true}}-g_{\text{predicted}}\|^2/\|g_{\text{true}}\|^2
    +\|g^{-1}_{\text{true}}-g^{-1}_{\text{predicted}}\|^2/\|g^{-1}_{\text{true}}\|^2
    +\|g^{-1}_{\text{predicted}}g_{\text{predicted}}-\Id_d\|^2, \nonumber
\end{align}
using Frobenius matrix norms. We train a neural network with two dense hidden layers to minimize \eqref{eq:approximator_loss}, and use this network for the geometry calculations. The network predicts the upper triangular part of each matrix, and this part is symmetrized to produce $g_{\text{predicted}}$ and $g^{-1}_{\text{predicted}}$. Note that additional methods could be employed to ensure the predicted metric being positive definite, see e.g. \cite{huang_riemannian_2016}. For the presented examples, it is our observations that the loss \eqref{eq:approximator_loss} ensures positive definiteness without further measures.

\section{Nonlinear Latent Space Statistics}
\label{sec:nonlinstat}
We now discuss aspects of nonlinear statistics applicable to the latent geometry setting. We start by focusing on means, particularly Fr\'echet and maximum likelihood (ML) means, before modeling variation around the mean with the principal geodesic analysis procedure.

\subsection{Fr\'echet and ML Means}
Fr\'echet mean \cite{frechet_les_1948} of a distribution on $M$ and its sample equivalent minimize the expected squared Riemannian distance: $\hat{x} = \argmin_{x\in M}\mathbb{E}[d(x,y)^2]$ and $\hat{x} = \argmin_{x\in M}\frac{1}{N}\sum_{i=1}^N d(x,y_i)^2$.
The standard way to estimate a sample Fr\'echet mean is to employ an iterative optimization to minimize the sum of squared Riemannian distances. For this, the Riemannian gradient of the squared distance can be expressed using the Riemannian Log map \cite{pennec_intrinsic_2006} by $\nabla_x d(x,y)^2  =  2\Log_x(y)$.

The Fr\'echet mean generalizes the Euclidean concept of a mean value as a distance minimizer. In Euclidean space, this is equivalent to the standard Euclidean estimator $\hat{x}=\frac{1}{N}\sum_i y_i$. From a probabilistic viewpoint, the equivalence between the log-density function of a Euclidean normal distribution and the squared distance results in $\hat{x}$ as an ML fit of a normal distribution to data:
\begin{equation}
  \hat{x}
  =
  \argmin_x\log p_{\mathcal N,x}(y),
  \label{eq:MLmean}
\end{equation}
with $p_{\mathcal N,x}(y)\propto\exp(-\frac12\|x-y\|^2)$ being the density of a normal distribution with mean $x$. While the normal distribution does not have a canonical equivalent on Riemannian manifolds, an intrinsic generalization comes from the transition density of a Riemannian Brownian motion. This density on $M$ arise as the solution to the heat PDE, $\frac{\partial}{\partial t}  p_{x,t}= \frac{1}{2} \Delta_g p_{x,t}$, using the Laplace-Beltrami operator $\Delta_g$, or, equivalently, from the law of the Brownian motion started at $M$. In \cite{sommer_modelling_2017,nye_construction_2014,sommer_bridge_2017}, this density is used to generalize the ML definition of the Euclidean mean
\begin{equation}
  \hat{x}
  =
  \argmin_x\log p_{x,T}(y),
  \label{eq:MLmean_manifold}
\end{equation}
for at fixed $T>0$.
We will develop approximation schemes for evaluating the log-density and for solving the optimization problem \eqref{eq:MLmean_manifold} in Section~\ref{sec:diffusions}.

\subsection{Principal Component Analysis}
Euclidean principal component analysis (PCA) estimates subspaces of the data space that explain the major variation of the data, either by maximizing variance or minimizing residuals. PCA is built around the linear vector space structure and the Euclidan inner product. Defining procedures that resemble PCA for manifold valued data hence become challenging, as neither inner products between arbitrary vectors nor the concept of linear subspaces is defined on manifolds.

Fletcher et al. \cite{fletcher_principal_2004-1} presented a generalized version of Euclidean PCA denoted principal geodesic analysis (PGA). PGA estimates nested geodesic submanifolds of $M$ that capture the major variation of the data projected to each submanifold. The geodesic subspaces hence take the place of the linear subspaces found with the Euclidean PCA.

Let $z_1,\ldots,z_N\in Z$ be latent space representations of the data $y_1,\ldots,y_N$ in $M$, and let $\mu$ be a Fr\'echet mean of the samples $z_1,\ldots,z_N$. We assume the observations are located in a neighbourhood $U$ of $\mu$ where $\Exp_\mu$ is invertible and the logarithm map, $\text{Log}_\mu$, thus well-defined. We then search for an orthonormal basis of tangent vectors in $T_\mu Z$ such that for each nested submanifold, $H_k = \text{Exp}_\mu(\text{span}\{v_1,\ldots,v_k\})$, the variance of the data projected on $H_k$ is maximized. The projection map used is based on the geodesic distance, $d$, and is defined by, $\pi_{H}(z) = \underset{z_1\in H}{\arg\min}\,d(z,z_1)^2$.

The tangent vectors $v_1,\ldots,v_k$ in the orthonormal basis of $T_\mu Z$ are found by optimizing the Fr\'echet variance of the projected data on the submanifold $H$, i.e.
\begin{align}
    v_k = \underset{\|v\|=1}{\arg\max}\sum_{i=1}^n d(\mu,\pi_{H}(z_i))^2,
\end{align}
where $H = \text{Exp}_\mu(\text{span}\{v_1,\ldots,v_{k-1},v\})$. For a more detailed description of the PGA procedure including computational approximations of the projection map in the tangent space of $\mu$ that we employ as well, see \cite{fletcher_principal_2004-1}. In the experiment section, we will perform PGA on the manifold defined by the latent space of a deep generative model for the MNIST dataset.

\section{Maximum Likelihood Inference of Latent Diffusions}
\label{sec:diffusions}
As in Euclidean statistics, parameters of distributions on manifolds can be inferred from data by maximum likelihood or, from a Bayesian viewpoint, maximum a posteriori. This can even be used to define statistical notions as exemplified by the ML mean in Section~\ref{sec:nonlinstat}. This probabilistic viewpoint relies on the existence of parametric families of distributions in the geometric spaces, and the ability to evaluate likelihoods. One example of such a distribution is the transition distribution of the Riemannian Brownian motion, see e.g. \cite{hsu_stochastic_2002}. In this section, we will show how likelihoods of data in the latent space $Z$ under this distribution can be evaluated by Monte Carlo sampling of conditioned diffusion bridges. As before, we assume the geometry of $Z$ has been trained by a separate training dataset, and that we wish to statistically analyze new observed data represented by $z_i$. To determine the transition distribution of a Brownian motion on the data manifold, we will apply a conditional diffusion bridge simulation procedure defined in~\cite{delyon_simulation_2006}, which will be described in the following section. This sampling scheme has previously been used for geometric spaces in \cite{arnaudon_geometric_2018,sommer_bridge_2017}.

\subsection{Bridge Simulation and Parameter Inference}
Let $z_1,\ldots,z_N\in Z$ be $N$ observations in $Z$. We assume $z_i$ are time $T$ observations from a Brownian motion, $z_t$ defined by $\eqref{eq:brown_motion}$, on $Z$ started at $x\in Z$. The aim is to optimize for the initial point $x$ by maximizing the likelihood of the observed data and thereby find the ML mean \eqref{eq:MLmean_manifold}. The mean value of the data distribution is thus defined as the starting point of the process maximizing the data likelihood, $L_\theta(z_1,\ldots,z_N) = \prod_{i=1}^N p_{T,\theta}(z_i)$, where $p_{T,\theta}(z_i)$ is the time $T$ transition density of $z_t$ evaluated at $z_i$.
The difficulty is to determine the transition density $p_{T,\theta}(z_i)$, i.e. the time $T$ density conditional on $z_T=z_i$. In~\cite{delyon_simulation_2006} it was shown that this conditional probability can be calculated based on the notion of a guided process
\begin{align}
    d\tilde{z}^j_t = g(z)^{kl}\Gamma^j_{kl}dt - \frac{\tilde{z}^j_t - z^j_i}{T-t}dt + \sqrt{(g(z))^{-1}}^jdB_t,
    \label{guided_proc}
\end{align}
which, without conditioning, almost surely hits the observation $z_i$ at time $t=T$. In fact, the guided process is absolutely continuous with respect to the conditional process $z_t|z_T=z_i$ with Radon-Nikodym derivative, $dP_{z|z_i}/dP_{\tilde{z}} = \varphi(\tilde{z})/E_{\tilde{z}}[\varphi(\tilde{z}_t)]$.
From this, the transition density can be expressed as
\begin{align}
    p_{T,\theta}(z_i) = \sqrt{\frac{|g(z_i)|}{(2\pi T)^{d}}}e^{-\frac{\|(x-z_i)^Tg(x)(x-z_i)\|^2}{2T}}\mathbb E_{\tilde{z}_t}[\varphi(\tilde{z}_t)],
\label{prob_bridge}
\end{align}
see~\cite{delyon_simulation_2006,sommer_bridge_2017} for more details. We can use Monte Carlo sampling of $\tilde{z}_t$ to approximate $\mathbb E_{\tilde{z}_t}[\varphi(\tilde{z}_t)]$ and hence determine $p_{T,\theta}(z_i)$ by \eqref{prob_bridge}. The likelihood can then be iteratively optimized to find the ML mean by computing gradients with respect to $x$.

\begin{figure}[htpb]
    \centering
    \subfigure{\includegraphics[width=.31\columnwidth,trim=170 120 130 150,clip=true]{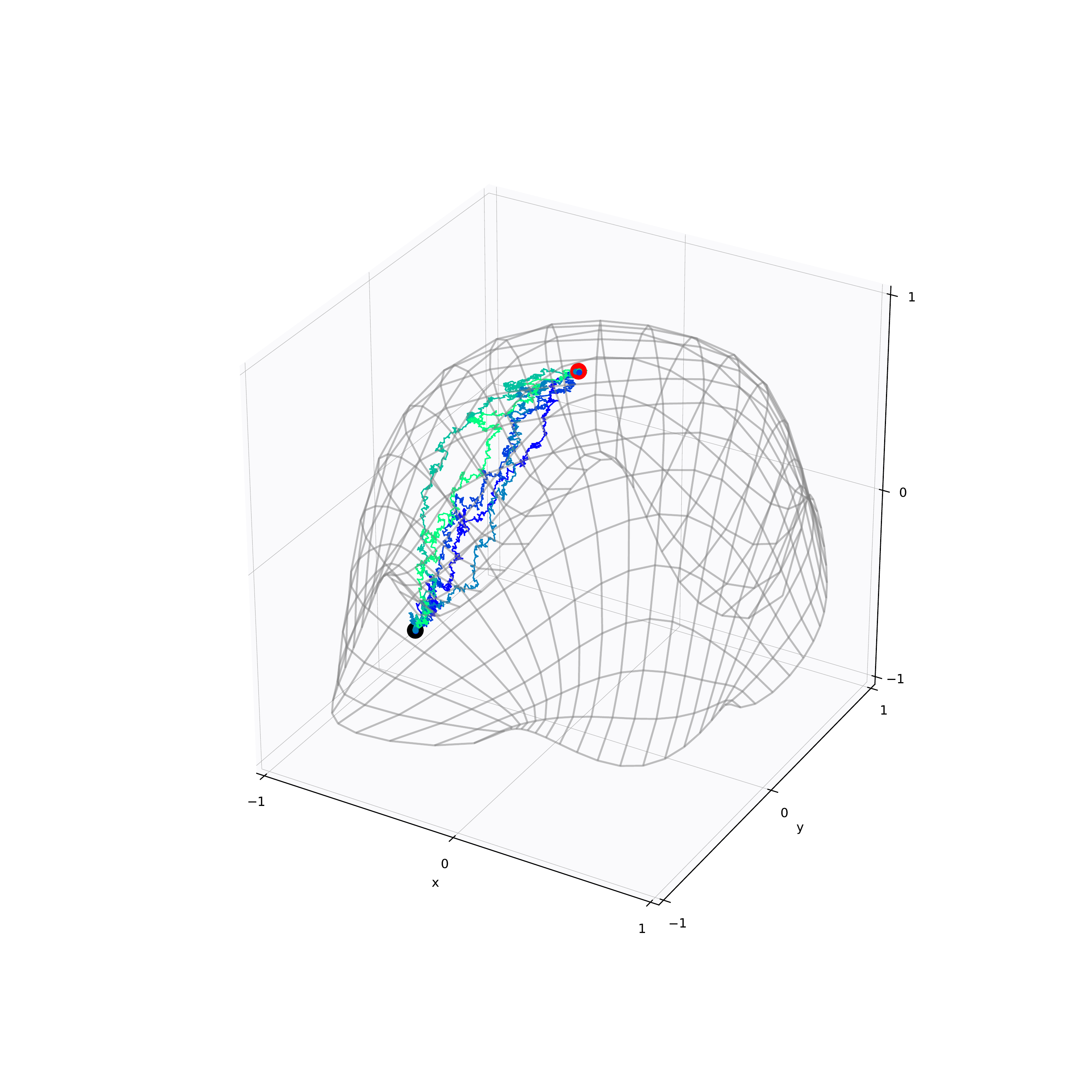}}
    \subfigure{\includegraphics[width=.31\columnwidth,trim=170 120 130 150,clip=true]{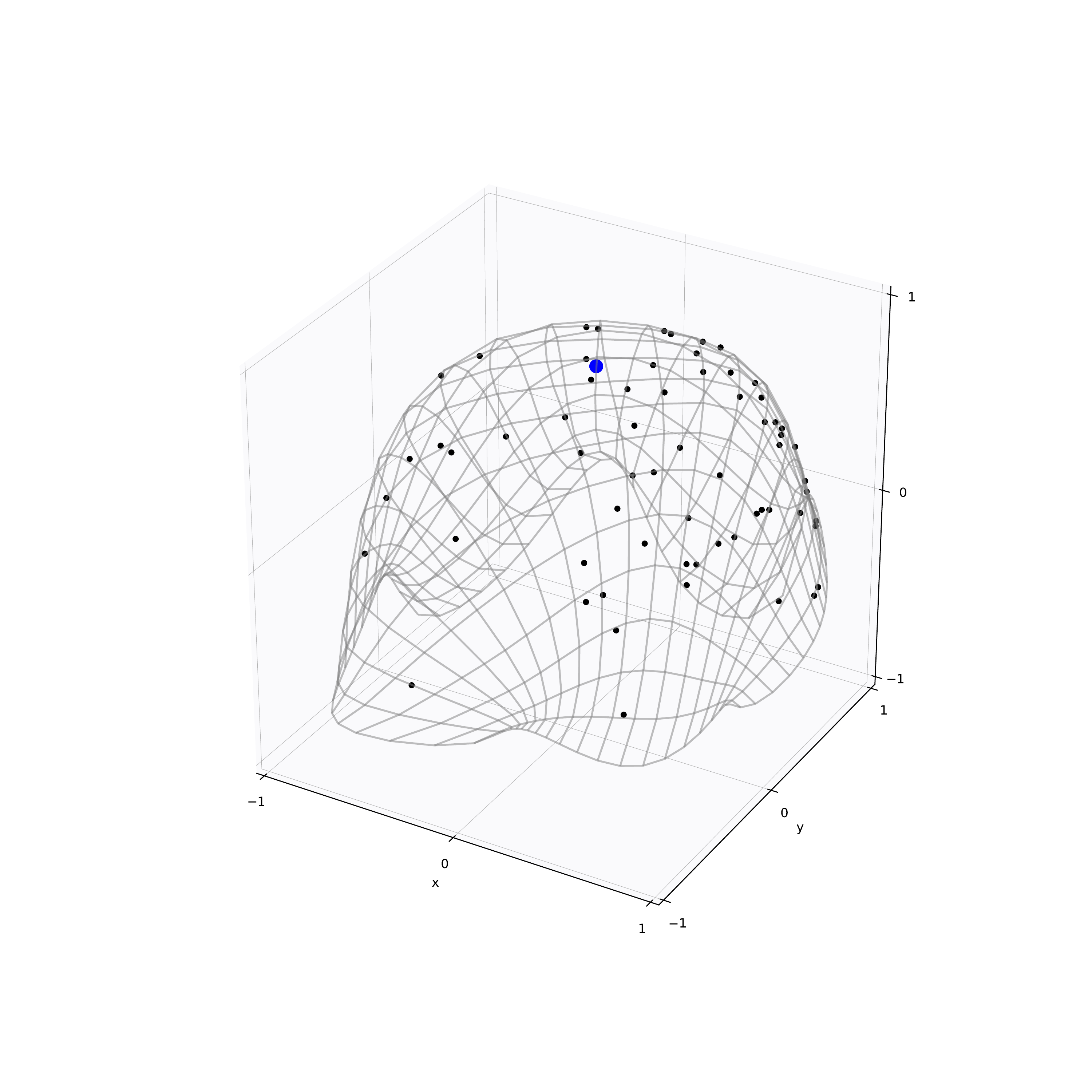}}
    \subfigure{\includegraphics[width=.31\columnwidth,trim=80 80 80 100,clip=true]{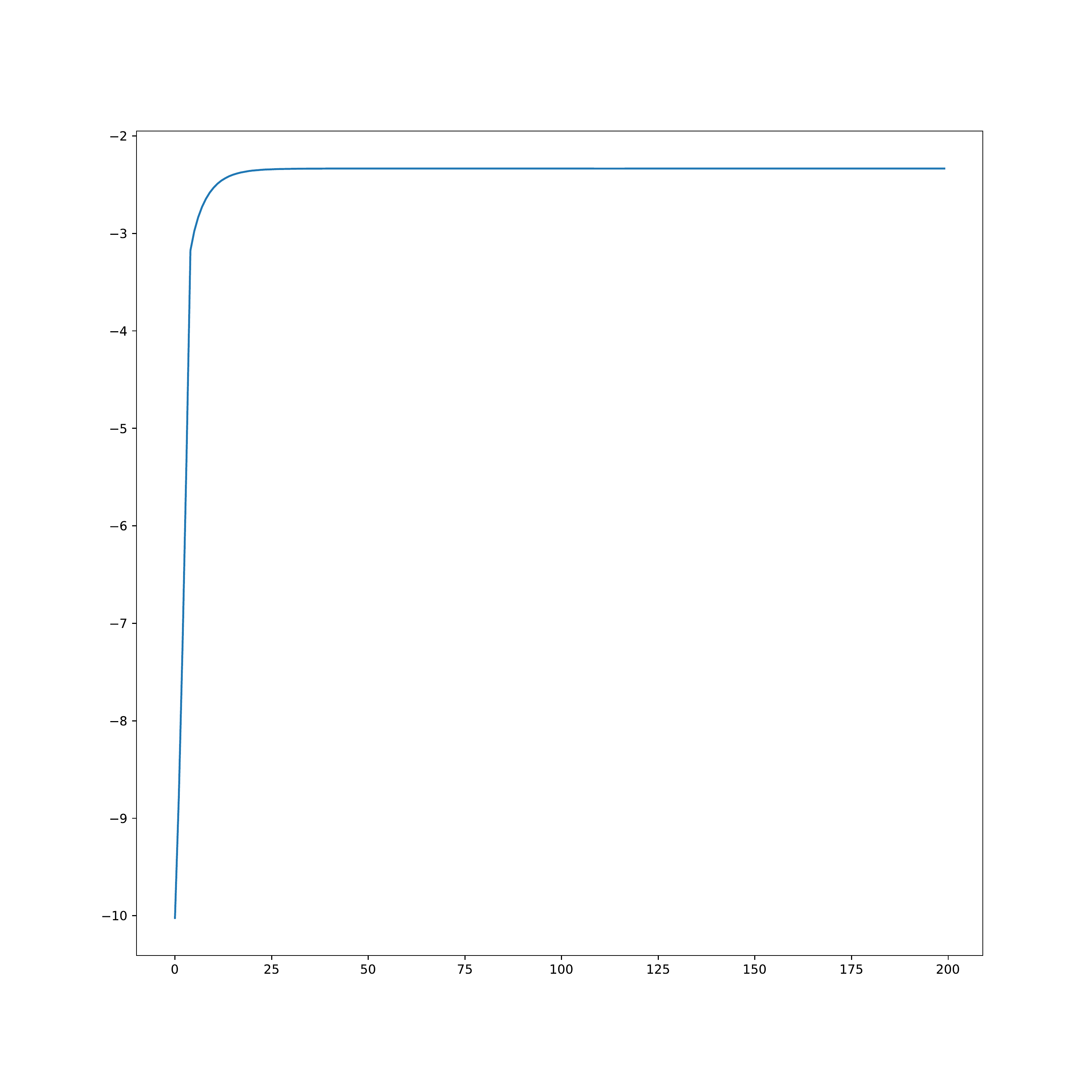}}
    \caption{(left) Brownian bridge sample paths on the trained data manifold. 
(middle) The estimated ML mean (blue) from the data (black points). 
 (right) The likelihood values from the MLE procedure.}
    \label{fig:S2_bridges}
\end{figure}

Figure \ref{fig:S2_bridges} shows sample paths of a Brownian bridge on the trained manifold for the synthetic data on $\mathbb{S}^2$ in addition to the ML estimated mean (middle). The likelihood values at each iteration are plotted in the same figure and illustrates that convergence has been reached for the MLE procedure.

\section{Experiments}

\label{sec:exp}
We will give examples of the analyses described above for the MNIST dataset~\cite{mnist}. The computations are performed with the Theano Geometry package \url{http://bitbucket.com/stefansommer/theanogeometry/} described in \cite{kuhnel_differential_2017}. The package contains implementations of differential geometry concepts and corresponding statistical algorithms.

\subsection{MNIST}

The MNIST dataset consists of images of handwritten digits from $0$ to $9$ with each observation of dimension $28\times 28$. A VAE has been trained on the full dataset providing a 2 dimensional latent space representation. The VAE \cite{kingma_auto-encoding_2013} has one hidden dense layer for both encoder and decoder, each layer containing $256$ neurons, and 2d latent space $Z$.

\begin{figure}[htpb]
    \centering
    \includegraphics[width=.312\columnwidth,trim=170 120 130 120,clip=true]{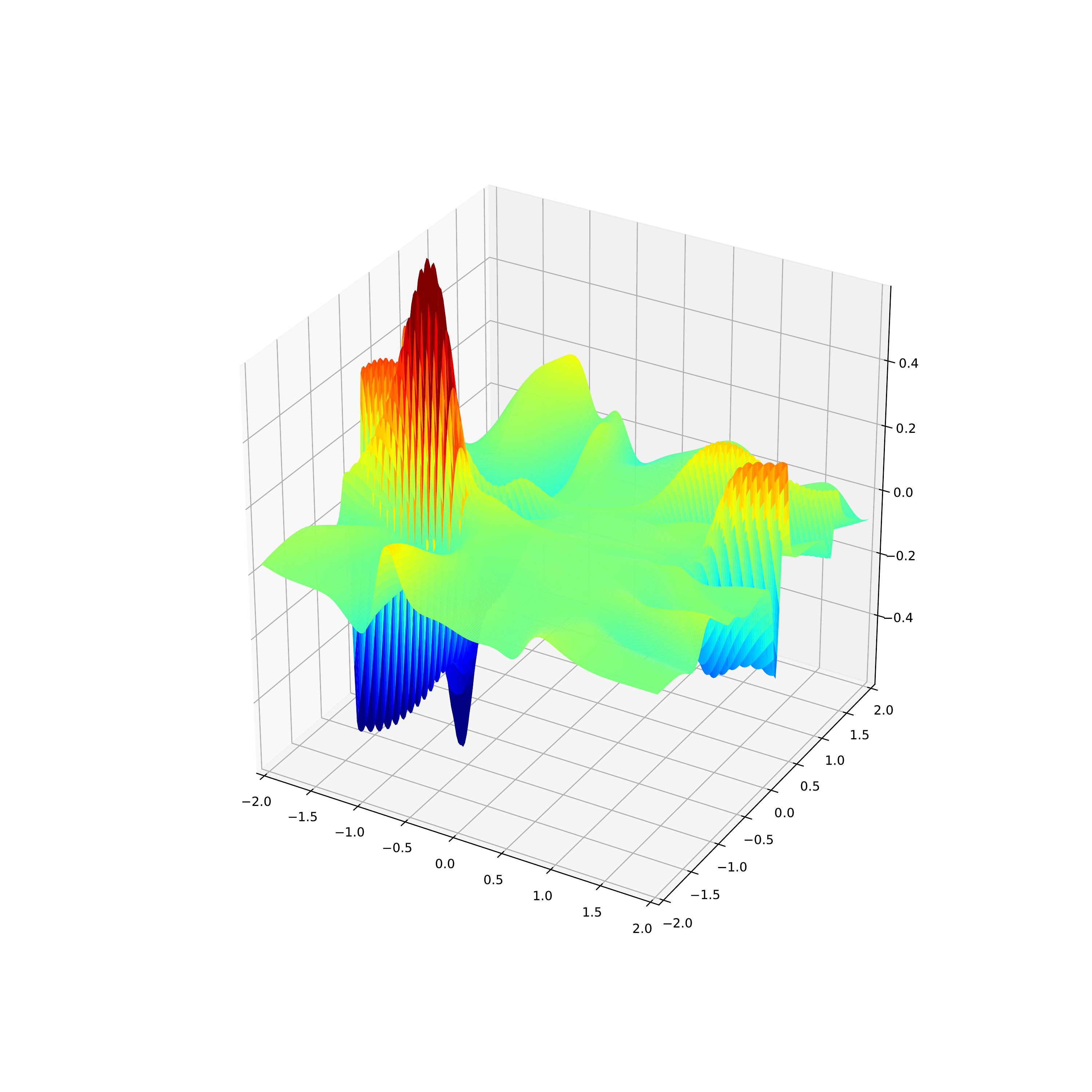}
    \includegraphics[width=.312\columnwidth,trim=170 120 130 120,clip=true]{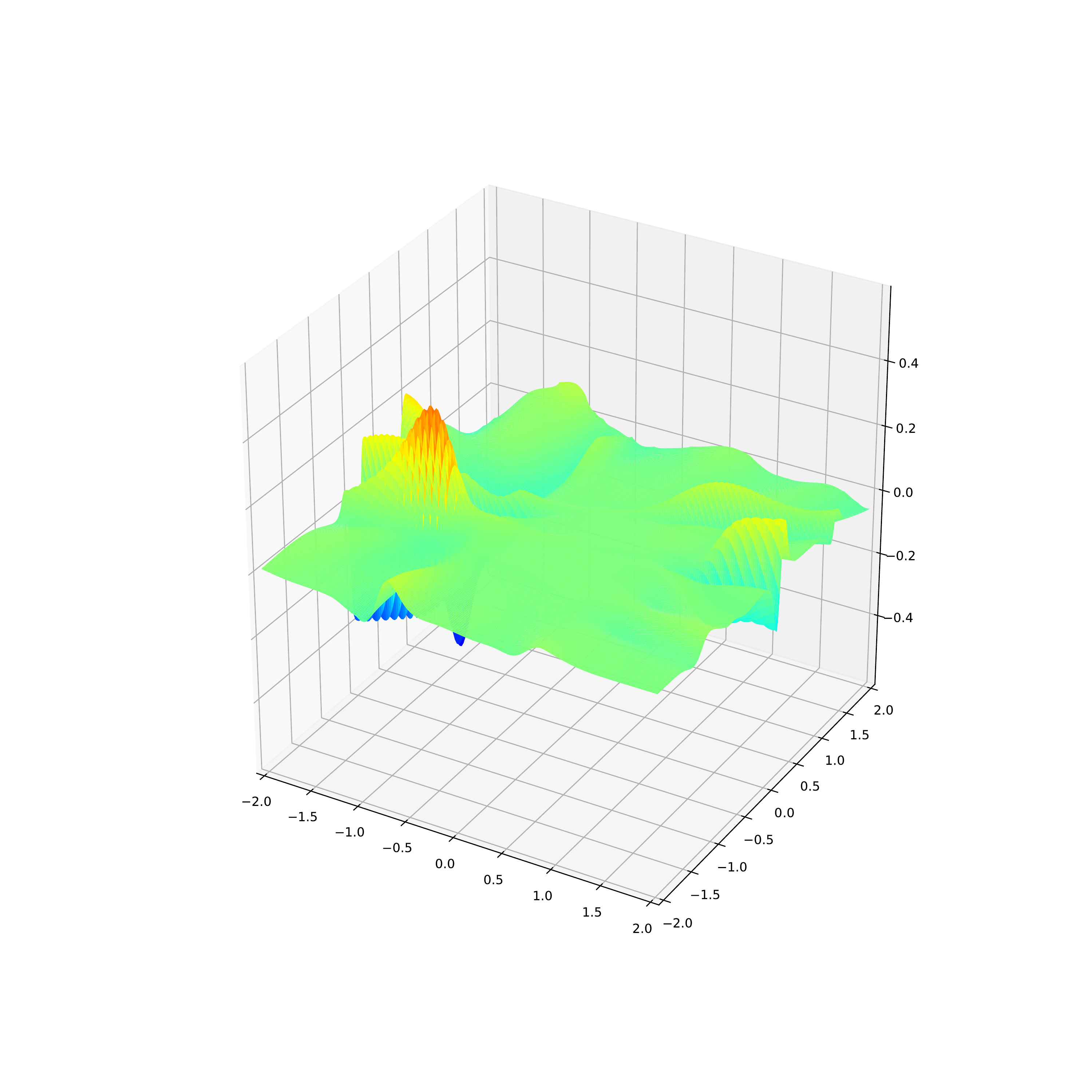}
    \includegraphics[width=.043\columnwidth,trim=670 100 100 100,clip=true]{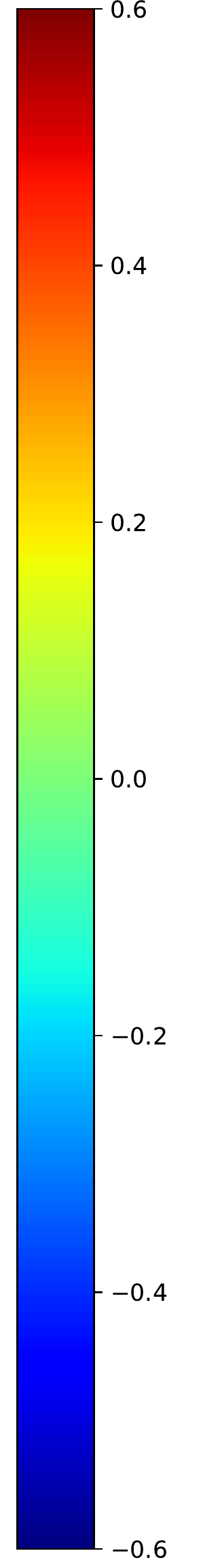}
    \includegraphics[width=.312\columnwidth,trim=80 80 80 80,clip=true]{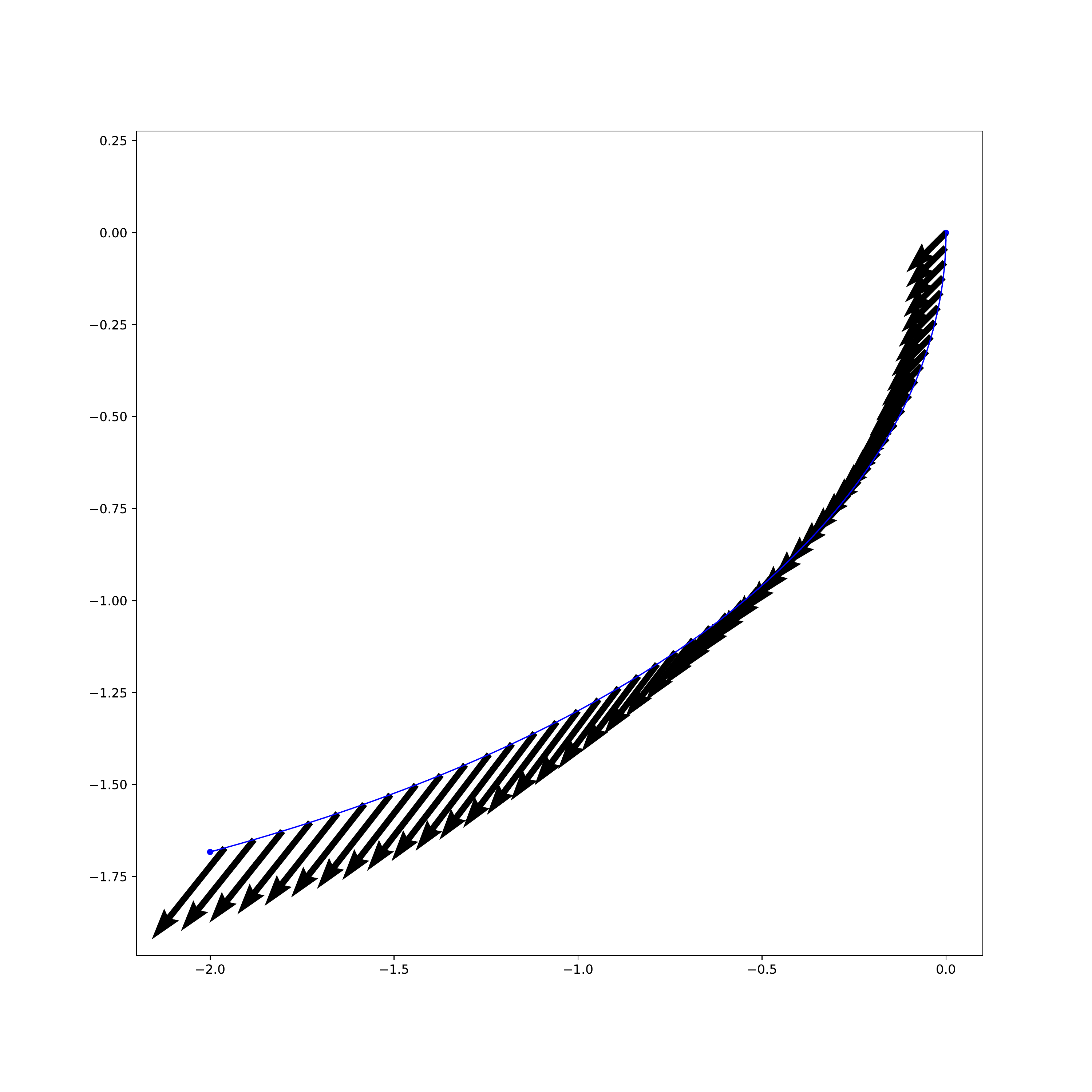}
    \caption{(left) Scalar curvature of space $Z$. (middle) Min. eigenvalue of the Ricci curvature tensor. (right) Parallel transport of a tangent vector in $Z$. The transported vectors have constant length measured by the Riemannian metric.}
    \label{fig:mnist_curvature}
\end{figure}

Figure \ref{fig:mnist_curvature} shows the scalar (left) and minimum Ricci curvature (middle) in a neighbourhood of the origin of $Z$. In addition, an example of parallel transport of a tangent vector along a curve in the latent space is visualized in the same figure (right). Note that the transported vector has constant length as measured by the metric $g$ which is clearly not the case for the Euclidean $\mathbb R^2$ norm.

\begin{figure}[htpb]
    \centering
    \subfigure{\includegraphics[width=.3\columnwidth,trim=80 80 80 100,clip=true]{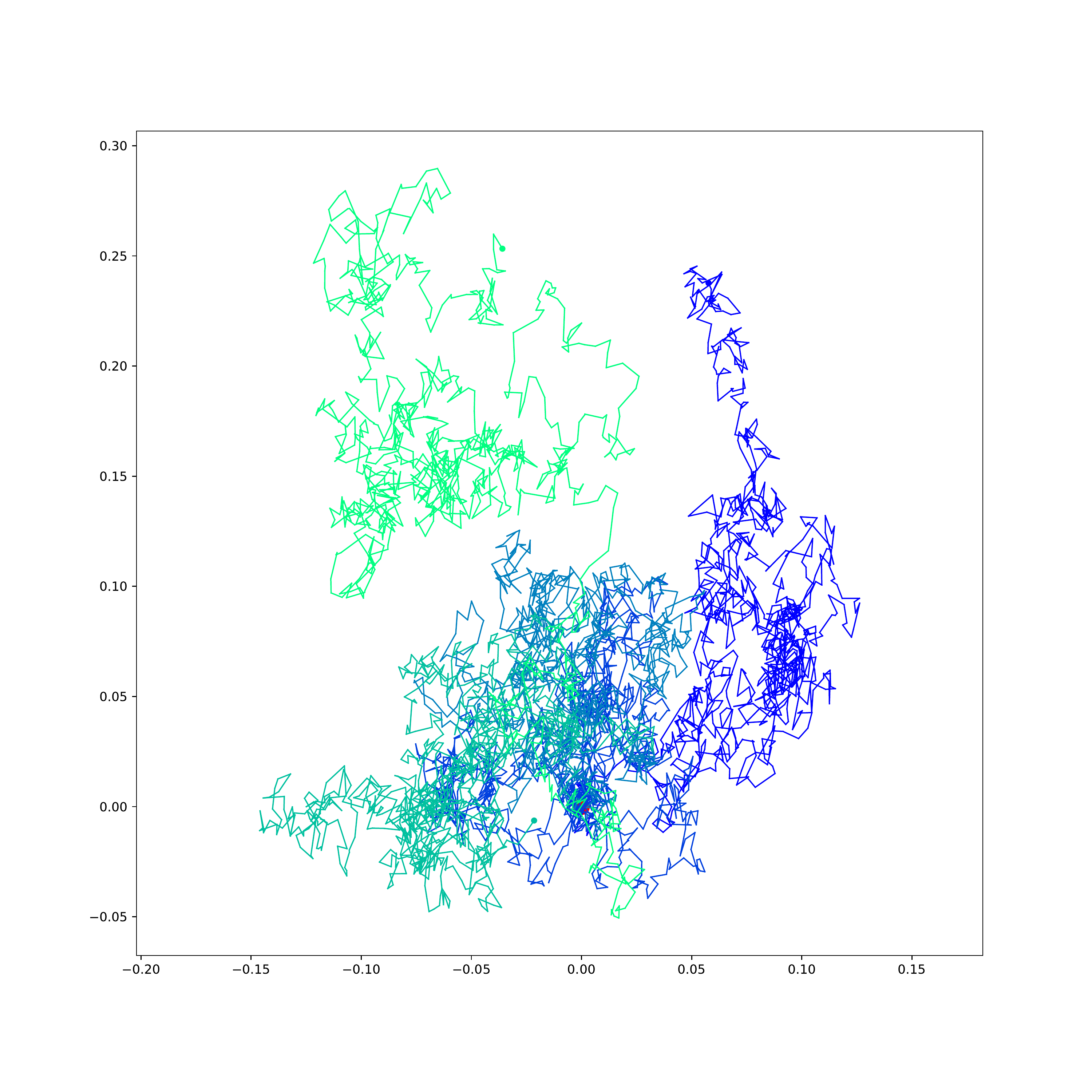}}\hspace{0.5cm}
    \subfigure{\includegraphics[width=.3\columnwidth,trim=80 80 80 100,clip=true]{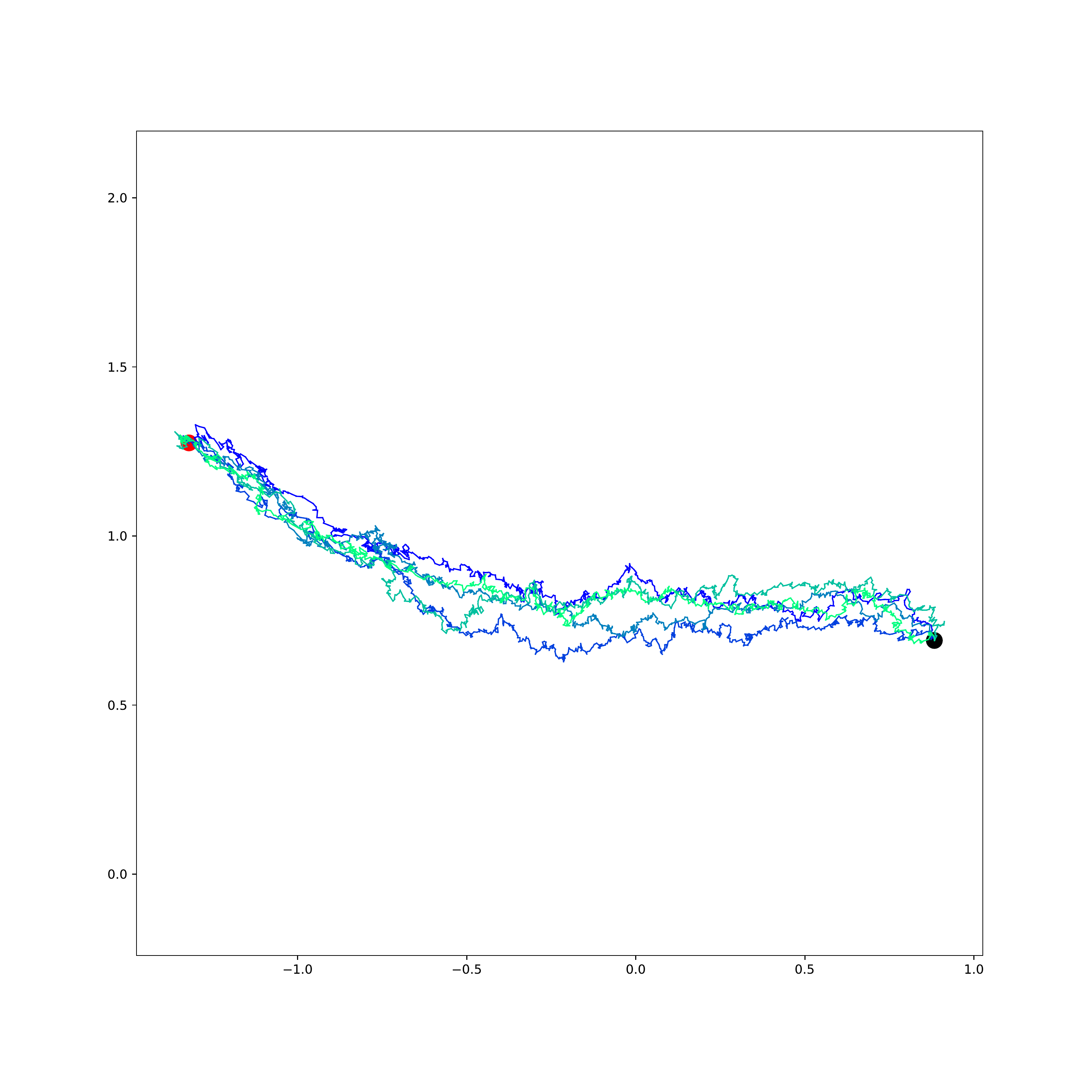}}
    \\ \vspace{-0.3cm}
    \begin{center}
        \includegraphics[width=1.\columnwidth,trim=100 85 90 85,clip=true]{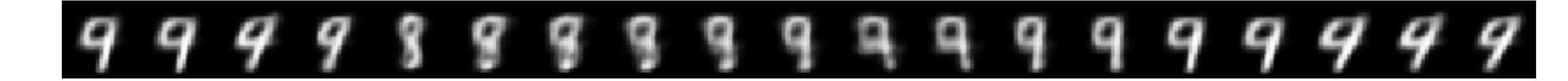}
        \includegraphics[width=1.\columnwidth,trim=100 85 90 85,clip=true]{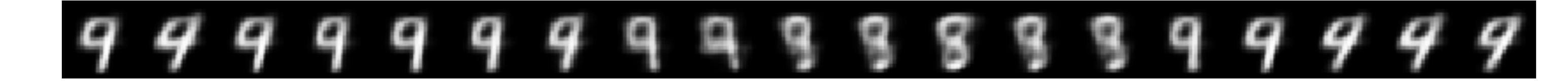}
        \includegraphics[width=1.\columnwidth,trim=100 85 90 85,clip=true]{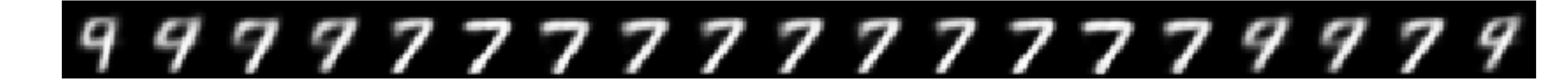}
    \end{center}\vspace{-0.2cm}
    \caption{(top left) Samples from a Riemannian Brownian motion in latent space. (top right) Samples from a Brownian bridge simulated by $\eqref{guided_proc}$. (bottom) Examples of Brownian bridges of MNIST data between two fixed 9s (left-/rightmost). The variance of the Brownian motion has been increased to visually emphasize the image variation.}
    \label{fig:mnist_stochastics}
\end{figure}

The top row of Figure \ref{fig:mnist_stochastics} shows samples of Brownian motions and Brownian bridges in the latent space $Z$. Each of these Brownian bridges correspond to a bridge in the data manifold of the MNIST data. Examples of bridges in the high dimensional space $X$ are shown in the bottom row of Figure \ref{fig:mnist_stochastics}. 

\begin{figure}[htpb]
    \centering
    \subfigure[Latent space representation]{\includegraphics[width=.32\columnwidth,trim=140 90 120 90,clip=true]{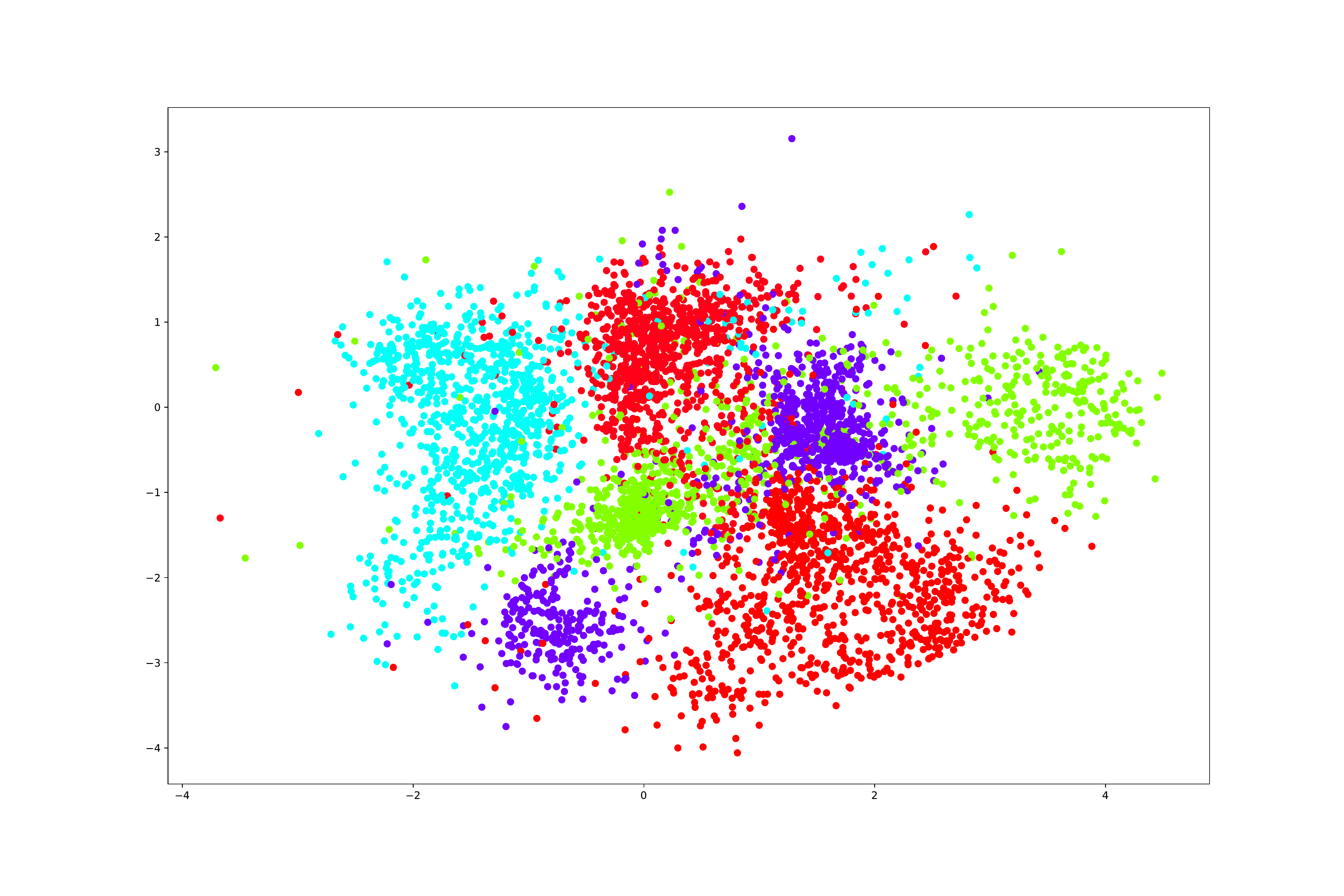}
    \label{fig:lat_rep}}
    \subfigure[PGA representation]{\includegraphics[width=.32\columnwidth,trim=140 90 120 90,clip=true]{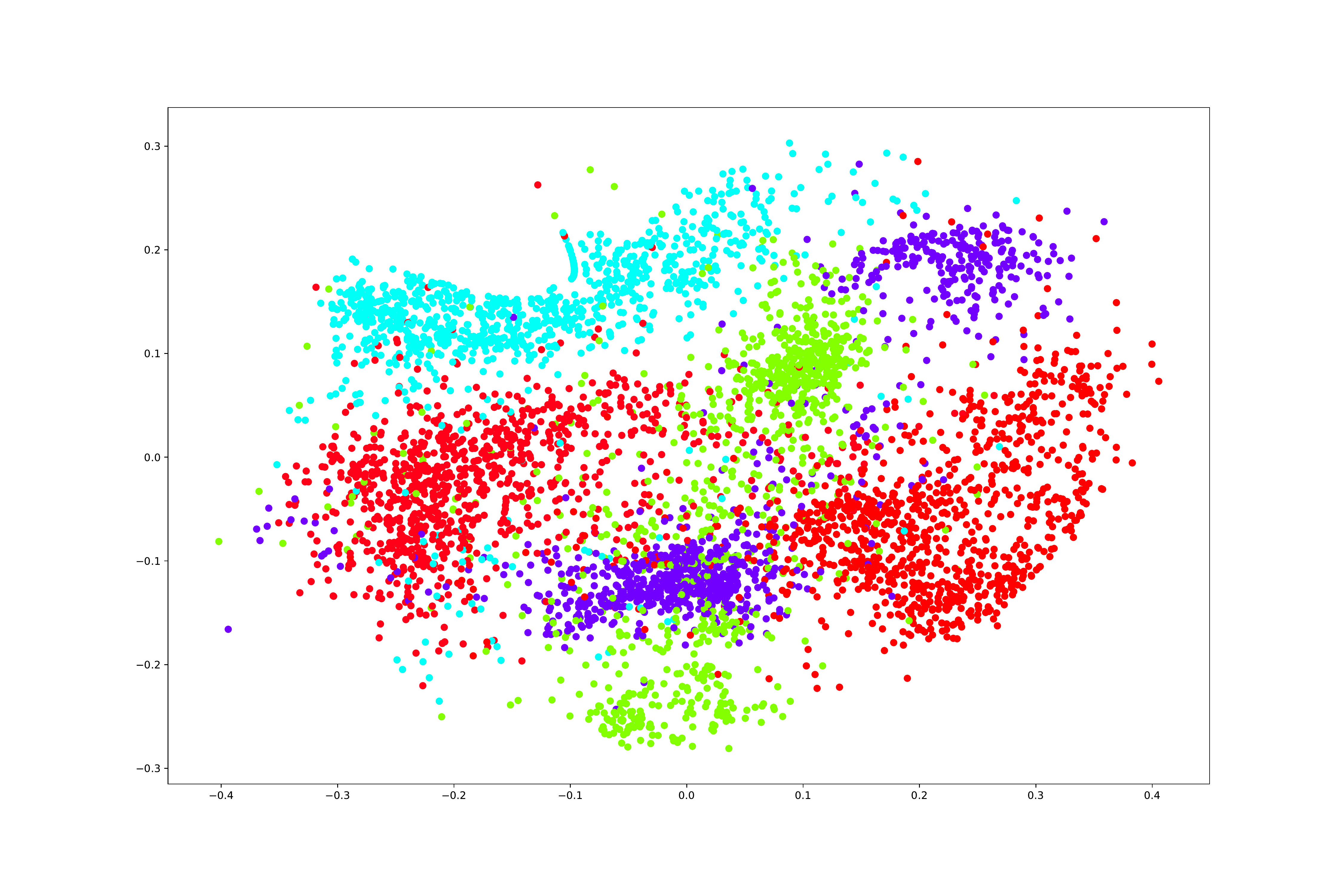}
    \label{fig:pga_rep}}
    \\
    \includegraphics[width=1.2\columnwidth,trim=100 85 -100 85,clip=true]{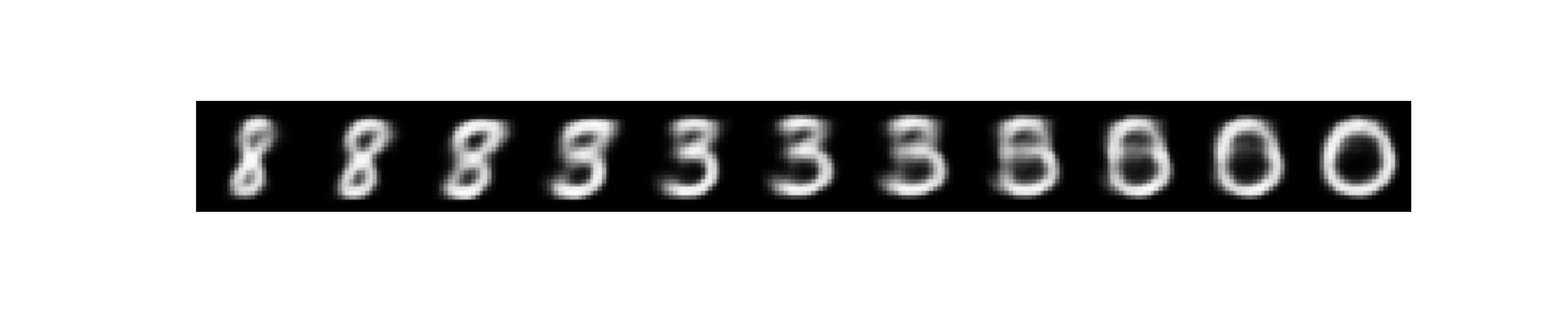}
    \includegraphics[width=1.2\columnwidth,trim=100 85 -100 85,clip=true]{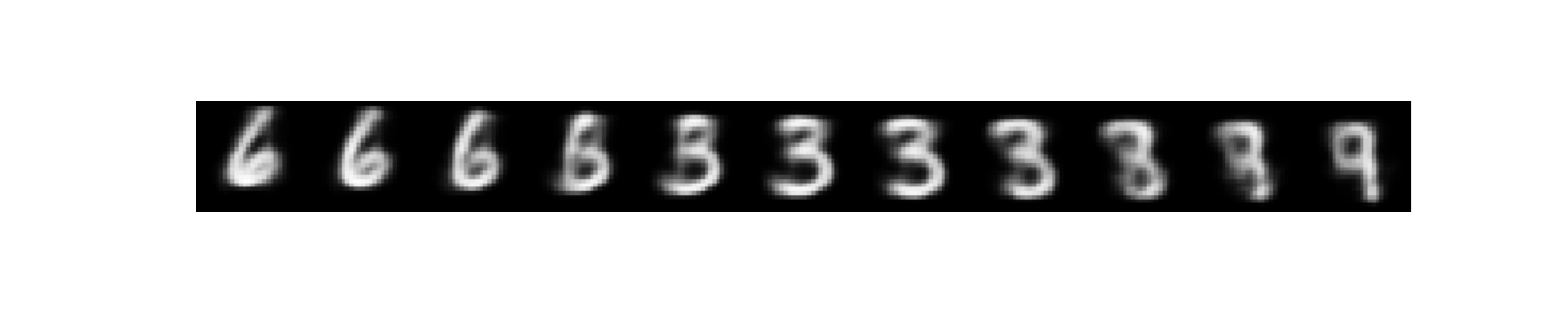}
    \caption{(top left) Latent space representation of data. (top right) PGA analysis on the sub-space of even digits. (bottom) Variation along first (1. row) and second (2. row) principal components.}
    \label{fig:mnist_pga}
\end{figure}

We now perform PGA on the latent space representation of the subset of the MNIST data consisting of even digits. PGA is a nonlinear coordinate change of the latent space around the Fr\'echet mean. PGA is applied to the data in Figure \ref{fig:lat_rep}, and the resulting transformed data in the PGA basis is shown in Figure \ref{fig:pga_rep}. The variation along the two principal component directions are visualized in the full dimensional data space in the bottom row of Figure \ref{fig:mnist_pga}.

\begin{figure}[htpb]
    \centering
    \includegraphics[width=0.245\columnwidth,trim=90 90 80 100,clip=true]{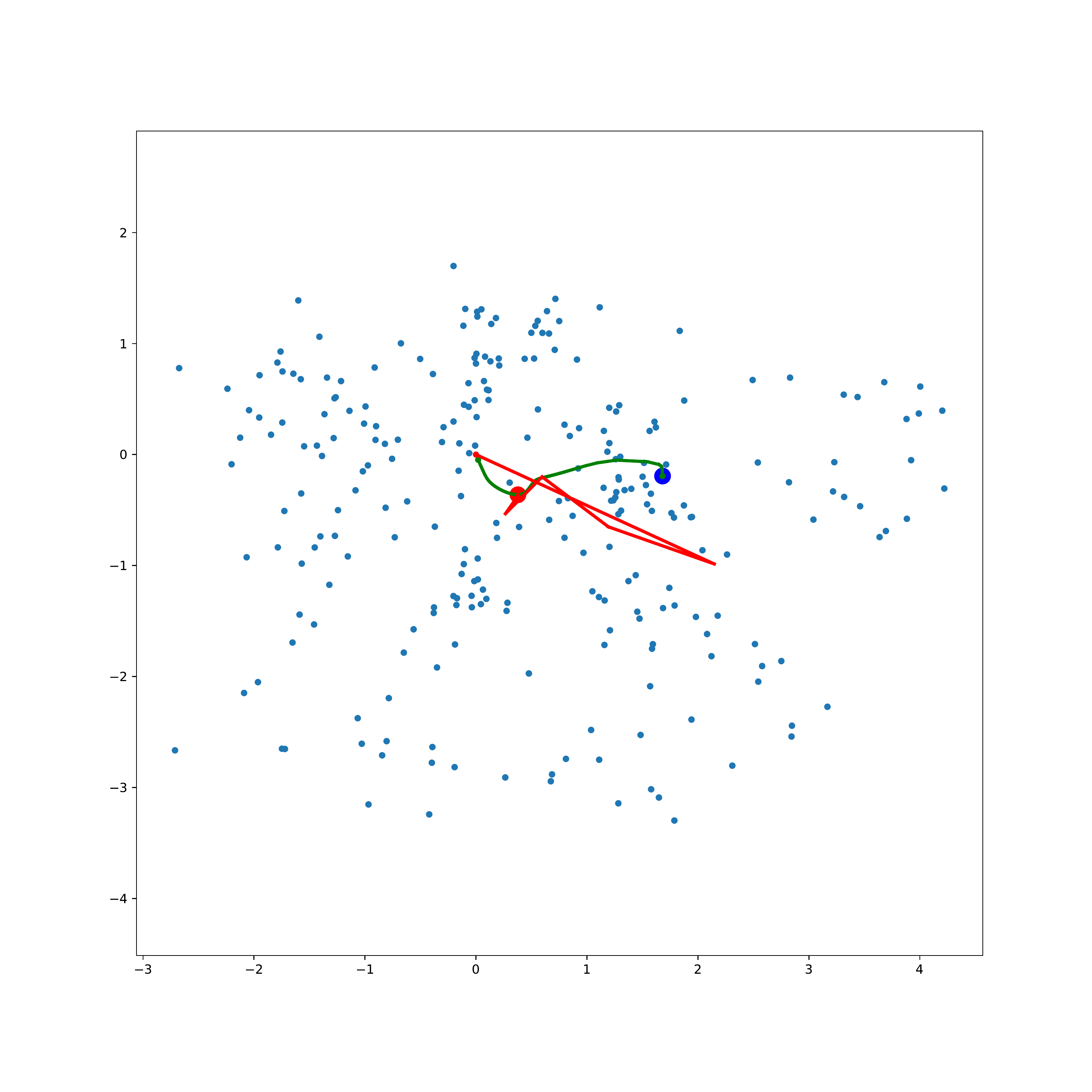}
    \includegraphics[width=0.245\columnwidth,trim=90 90 80 100,clip=true]{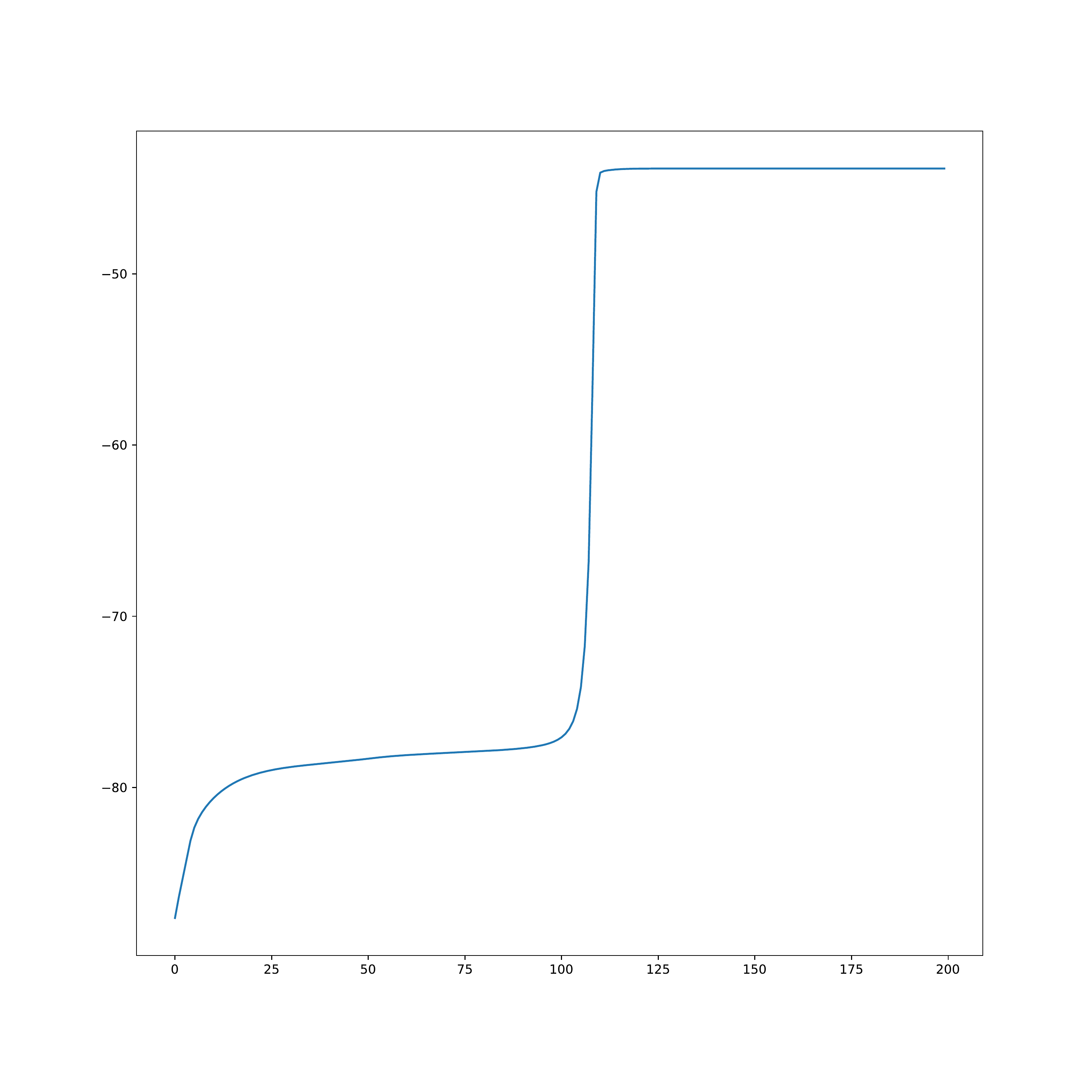}
    \includegraphics[width=0.245\columnwidth,trim=90 90 90 100,clip=true]{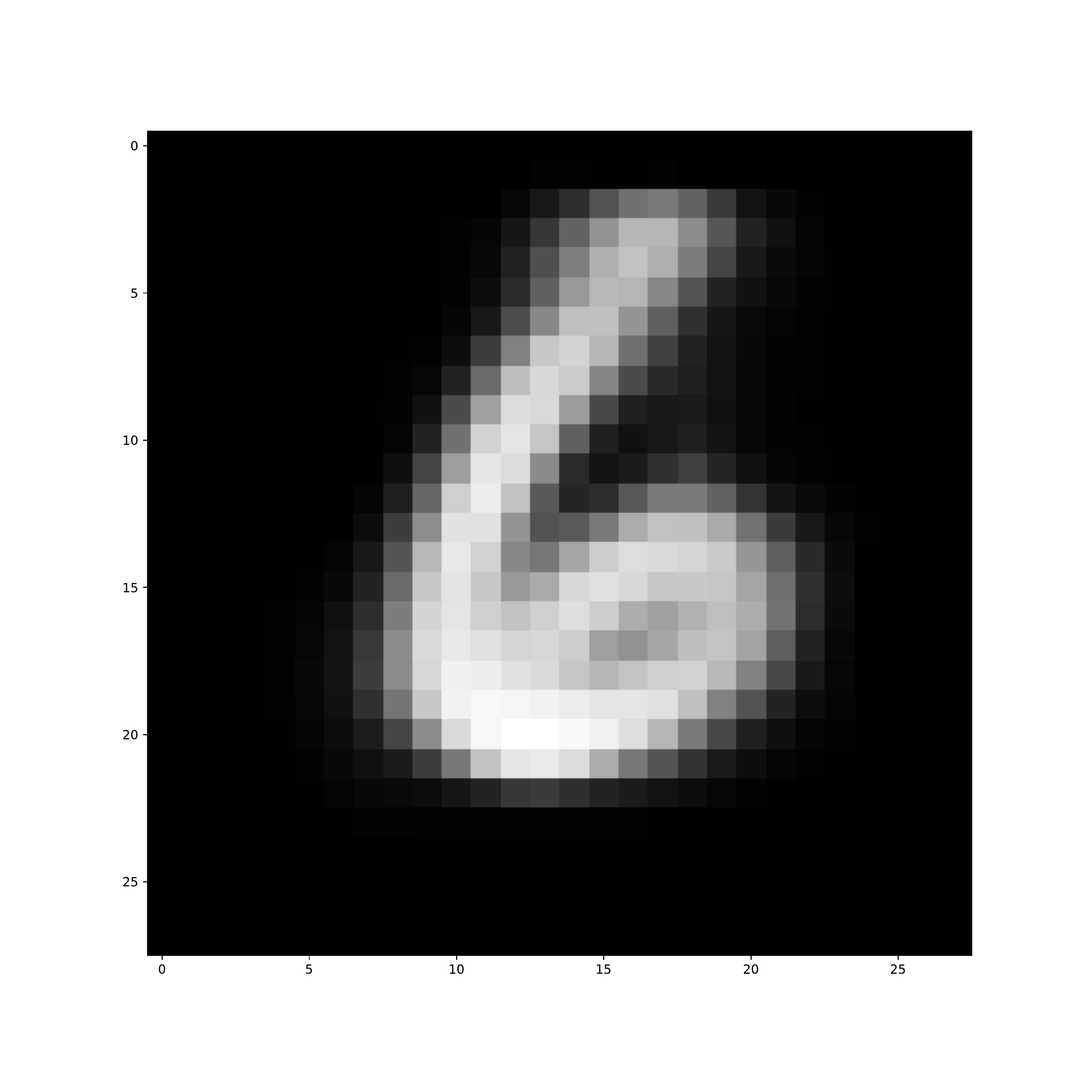}
    \includegraphics[width=0.245\columnwidth,trim=90 90 90 100,clip=true]{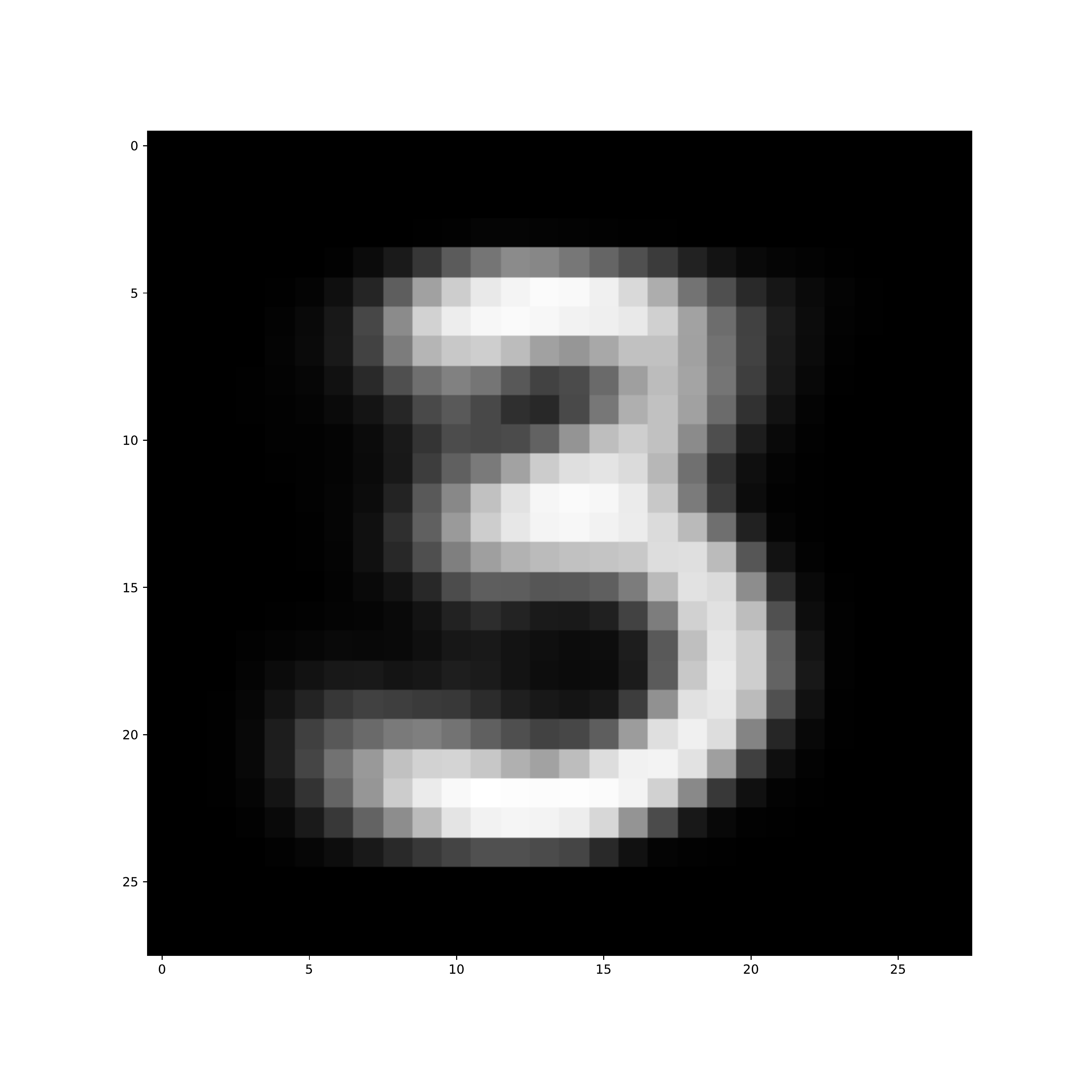}
    \caption{From left: (1.) Iterations of ML (green) and Fr\'echet mean (red) for subset of even MNIST digits. (2.) Likelihood evolution during the MLE. Estimated ML mean (3.) and Fr\'echet mean (4.).}
    \label{fig:mnist_ML}
\end{figure}

Figure \ref{fig:mnist_ML} (3. image from left) shows the maximum likelihood mean image for a subset of 256 even digits estimated by the ML procedure described in section \ref{sec:diffusions}. Figure \ref{fig:mnist_ML} (right) shows the corresponding Fr\'echet mean. The iterations, for both ML and Fr\'echet mean, in latent space are shown in Figure \ref{fig:mnist_ML} (left), with the last plot showing the likelihood values for each step of the ML optimization.

\section{Conclusion}
Deep generative models define an embedding of a low dimensional latent space $Z$ to a high dimensional data space $X$. The embedding can be used to reduce data dimensionality and move statistical analysis from $X$ to the low-dimensional latent representation in $Z$. This method can be seen as a nonlinear equivalent to the dimensionality reduction commonly performed by PCA. Nonlinear structure in data can be represented compactly, and the induced geometry necessitates use of nonlinear statistical tools. We considered principal geodesic analysis on the latent space and maximum likelihood estimation of the mean using simulation of conditioned diffusion processes. To enable fast computation of the geometric algorithms that involve high-order derivatives of the metric, we fit a second neural network, to predict the metric $g$ and its inverse, which vastly speeds up computations. We visualized examples on 3D synthetic data simulated on $\mathbb{S}^2$ and performed analyses on the MNIST dataset based on a trained VAE with a 2D latent space.

\bibliographystyle{plain}
\bibliography{ss.bib}

\end{document}